\documentclass{llncs}

\usepackage{amsmath,amssymb,graphicx}
\usepackage{pstricks}
\usepackage[a4paper,hmargin=1in,vmargin=1.7in]{geometry}
\usepackage{epstopdf}
\usepackage{epsfig}

\newcommand{\N}{{\mathbb N}}

\newcommand{\R}{{\mathbb R}}

\newcommand{\ud}{{\mathrm d}}       
\newcommand{\overoverline}[1]{\overline{\overline{#1}}}

\newcommand{\xs}{{x^*}} 

\title{{Learning Low-Density Separators}}
\author{Shai Ben-David\inst{1} \and Tyler Lu\inst{1} \and D\'{a}vid P\'{a}l\inst{1} \and Miroslava Sot\'{a}kov\'{a}\inst{2}}

\institute{
David R. Cheriton School of Computer Science\\
University of Waterloo, ON, Canada\\
\email{\{shai,ttlu,dpal\}@cs.uwaterloo.ca}
\and
Department of Computer Science\\
University of Aarhus, Denmark\\
\email{mirka@cs.au.dk}
}

\begin{document}

\maketitle

\begin{abstract}
We define a novel, basic, unsupervised learning problem - learning
the lowest density homogeneous hyperplane separator of an
unknown probability distribution.  This task is relevant to several
problems in machine learning, such as semi-supervised learning and
clustering stability.  We investigate the question of existence of a
universally consistent algorithm for this problem. We propose two
natural learning paradigms and prove that, on input unlabeled random
samples generated by any member of a rich family of distributions,
they are guaranteed to converge to the optimal separator for that
distribution. We complement this result by showing that no learning
algorithm for our task can achieve uniform learning rates (that are
independent of the data generating distribution).
\end{abstract}

\section{Introduction}
While the theory of machine learning has achieved extensive
understanding of many aspects of supervised learning, our
theoretical understanding of unsupervised learning leaves a lot to
be desired. In spite of the obvious practical importance of various
unsupervised learning tasks, the state of our current knowledge does
not provide anything that comes close to the rigorous mathematical
performance guarantees that classification prediction theory enjoys.

In this paper we make a small step in that direction by
analyzing one specific unsupervised learning
task -- the detection of low-density linear separators for data
distributions over Euclidean spaces.

We consider the following task: \emph{for an unknown data
distribution over $\R^n$, find the homogeneous hyperplane of
lowest density that cuts through that distribution.} We assume that
the underlying data distribution has a continuous density function
and that the data available to the learner are finite i.i.d. samples
of that distribution.

Our model can be viewed as a restricted instance of the fundamental
issue of inferring information about a probability distribution from
the random samples it generates. Tasks of that nature range from the
ambitious problem of density estimation \cite{DevroyeL01}, through
estimation of level sets \cite{BDL97}, \cite{Tsybakov1997},
\cite{SSN07}, densest region detection \cite{BDES02}, and, of course,
clustering. All of these tasks are notoriously difficult with
respect to both the sample complexity and the computational
complexity aspects (unless one presumes strong restrictions about
the nature of the underlying data distribution). Our task seems more
modest than these. Although we are not aware of any previous work on
this problem (from the point of view of statistical machine
learning, at least), we believe that it is a rather basic problem
that is relevant to various practical learning scenarios.

One important domain to which the detection of low-density linear
data separators is relevant is semi-supervised learning
\cite{ChaSchZie06}. Semi-supervised learning is motivated by the
fact that in many real world classification problems, unlabeled
samples are much cheaper and easier to obtain than labeled examples.
Consequently, there is great incentive to develop tools by which
such unlabeled samples can be utilized to improve the quality of
sample based classifiers. Naturally, the utility of unlabeled data
to classification depends on assuming some relationship between the
unlabeled data distribution and the class membership of data points
 (see \cite{BDLP08} for a rigorous discussion of this point).
A common postulate of that type is that the boundary
between data classes passes through low-density regions of the data
distribution. The Transductive Support Vector Machines paradigm
(TSVM)~\cite{Joachims99} is an example of an algorithm that
implicitly uses such a low density boundary assumption. Roughly
speaking, TSVM searches for a hyperplane that has small error on the
labeled data and at the same time has wide margin with respect to
the unlabeled data sample.

Another area in which low-density boundaries play a significant role
is the analysis of clustering stability. Recent work on the analysis
of clustering stability found close relationship between the
stability of a clustering and the data density along the cluster
boundaries -- roughly speaking, the lower these densities the more stable the
clustering (\cite{BDL08}, \cite{ST08}).

A low-density-cut algorithm for a family $\mathcal{F}$ of probability distributions
takes as an input a finite sample generated by some distribution $f\in\mathcal{F}$
and has to output a hyperplane through the origin with low density w.r.t. $f$.
In particular, we consider the family of all distributions over $\R^n$ that
have continuous density functions.
We investigate two notions of success for
low-density-cut algorithms -- uniform convergence (over a family of
probability distributions) and consistency. For uniform convergence
we prove a general negative result, showing that no algorithm can
guarantee any fixed convergence rates (in terms of sample sizes).
This negative result holds even in the simplest case where the data
domain is the one-dimensional unit interval. For consistency (e.g.,
allowing the learning/convergence rates to depend on the
data-generating distribution), we prove the success of two natural
algorithmic paradigms; \emph{Soft-Margin} algorithms that choose a
margin parameter (depending on the sample size) and output the
separator with lowest empirical weight in the margins around it, and
\emph{Hard-Margin} algorithms that choose the separator with widest
sample-free margins.

The paper is organized as follows: Section \ref{preliminaries}
provides the formal definition of our learning task as well as the
success criteria that we investigate. In Section
\ref{section:onedim} we present two natural learning paradigms for
the problem over the real line and prove their universal consistency
over a rich class of probability distributions. Section
\ref{section:highdim} extends these results to show the learnability
of lowest-density homogeneous linear cuts for probability
distributions over $R^d$ for arbitrary dimension, $d$. In Section
\ref{section:no-uniform-convergence} we show that the previous
universal consistency results cannot be improved to obtain
\emph{uniform} learning rates (by any finite-sample based
algorithm). We conclude the paper with a discussion of directions
for further research.

\section{Preliminaries}
\label{preliminaries}

We consider probability distributions over $\R^d$.
For concreteness, let the domain of the
distribution be the $d$-dimensional unit ball.

A \emph{linear cut learning algorithm} is an algorithm that takes as
input a finite set of domain points, a sample $S \subseteq \R^d$,
and outputs a homogenous hyperplane, $L(S)$ (determined by a weight
vector, $\mathbf{w} \in \R^d$, such that $||\mathbf{w}||_2=1$).

We investigate algorithms that aim to detect hyperplanes with low
density with respect to the sample-generating  probability
distribution.

Let $f:\R^d \to \R_0^+$ be a $d$-dimensional density function. We
assume that $f$ is continuous. For any homogeneous hyperplane
$h(\mathbf{w})=\{ \mathbf{x} \in \R^d ~:~ \mathbf{w}^T \mathbf{x} =
0 \}$ defined by a unit weight vector $\mathbf{w} \in \R^d$, we
consider the $(d-1)$-dimensional integral of the density over $h$,
$$
\overline{f}(\mathbf{w}) := \int_{h(\mathbf{w})} f(\mathbf{x}) \; \ud x \; .
$$

Note that $\mathbf{w} \mapsto \overline{f}(\mathbf{w})$ is a
continuous mapping defined on the $(d-1)$-sphere $\mathcal{S}^{d-1}
= \{ \mathbf{w} \in \R^d ~:~ \|\mathbf{w}\|_2 = 1 \}$.  Note that,
for any such weight vector $\mathbf{w}$,  $\overline{f}(\mathbf{w})
= \overline{f}(-\mathbf{w})$. For the 1-dimensional case, these
hyperplanes are replaced by points, $\mathbf{x}$ on the real line,
and $\overline{f}(\mathbf{x}) = f(\mathbf{x})$ -- the density at the
point $\mathbf{x}$.

\begin{definition}
A {\em linear cut learning algorithm} is a function that maps
samples to homogeneous hyperplanes. Namely,
$$
L:\bigcup_{m=1}^\infty (\R^d)^m \to \mathcal{S}^{d-1}.
$$

When $d=1$, we require that
$$
L:\bigcup_{m=1}^\infty \R^m \to [0,1].
$$

(The intention is that $L$ finds the lowest density linear separator
of the sample generating distribution.)

\end{definition}

\begin{definition}
Let $\mu$ be a probability distribution and $f$ its density
function. For a weight vector $\mathbf{w}$ we define the half-spaces
$h^+(\mathbf{w}) = \{ \mathbf{x} \in \R^d ~:~ \mathbf{w}^T
\mathbf{x} \ge 0 \}$ and $h^-(\mathbf{w}) = \{ \mathbf{x} \in \R^d
~:~ \mathbf{w}^T \mathbf{x} \le 0 \}$. For any weight vectors
$\mathbf{w}$ and $\mathbf{w}'$,
\begin{enumerate}
\item $D_E(\mathbf{w,w'})=1-|\mathbf{w^Tw}'|$
\item $D_{\mu}(\mathbf{w},\mathbf{w}') = \min\{\mu(h^+({\mathbf{w}}) \Delta
h^+({\mathbf{w}'})), \mu(h^-({\mathbf{w}}) \Delta
h^+({\mathbf{w}'}))\}$
\item $D_{f}(\mathbf{w},\mathbf{w}') = | \overline{f}(\mathbf{w}') -
\overline{f}(\mathbf{w}) |$
\end{enumerate}
\end{definition}

We shall mostly consider the distance measure $D_E$ in $\R^d$, for
$d>1$ and $D_E(x,y)=|x-y|$ for $x,y \in \R$. In theses cases we omit
any explicit reference to $D$. All of our results hold as well when
$D$ is taken to be the probability mass of the symmetric difference
between $L(S)$ and $\mathbf{w}^*$ and when $D$ is taken to be
$D(\mathbf{w, w'})=| \overline{f}(\mathbf{w})-
\overline{f}(\mathbf{w'})|$.

\begin{definition}
Let ${\cal F}$ denote a family of probability distributions over $\R^d$.
We assume that all members of ${\cal F}$
have density functions, and identify a distribution with its density
function. Let $D$ denote a distance function over hyperplanes. For a
linear cut learning algorithm, $L$, as above,
\begin{enumerate}
\item[1.] We say that $L$, is \emph{consistent} for ${\cal F}$ w.r.t a distance
measure $D$, if, for any probability distribution $f$ in ${\cal F}$,
if $f$ attains a unique minimum density hyperplane then
\begin{equation}
\label{equation:consistency-hyperplanes} \forall \epsilon > 0 \qquad
\lim_{m \to \infty} \Pr_{S \sim f^m} \left[ D(L(S), \mathbf{w}^*)
\ge \epsilon \right] = 0 \; .
\end{equation}
where $\mathbf{w}^*$ is the minimum density hyperplane for $f$.
\end{enumerate}

\begin{enumerate}
\item[2.] We say that $L$ is \emph{uniformly convergent} for
${\cal F}$ (w.r.t a distance measure, $D$), if, for every
$\epsilon,\delta>0$, there exists a $m(\epsilon,\delta)$ such that
for any probability distribution $f \in {\cal F}$,  if $f$ has a
unique minimizer $\mathbf{w}^*$ then, for all $m\geq
m(\epsilon,\delta)$ we have
\begin{equation} \label{eqn:uniform_convergence}
\Pr_{S\sim f^m} \left[ D(L(S), \mathbf{w^*}) \ge \epsilon \right]
\leq \delta.
\end{equation}
\end{enumerate}
\end{definition}


\section{The One Dimensional Problem}\label{section:onedim}
%
Let $\mathcal{F}_1$ be the family of all probability distributions
over the unit interval $[0,1]$ that have continuous density
function. We consider two natural algorithms for lowest density cut
over this family. The first is a simple bucketing algorithm. We
explain it in detail and show its consistency in
section~\ref{section:bucketing}. The second algorithm is the
\emph{hard-margin} algorithm which outputs the mid-point of the
largest gap between two consecutive points the sample. In
section~\ref{section:hard-margin} we show \emph{hard-margin}
algorithm is consistent and in section~\ref{section:bucketing} that
the bucketing algorithm is consistent. In section
\ref{section:no-uniform-convergence} we show there are no algorithms
that are uniformly convergent for $\mathcal{F}_1$.

\subsection{The Bucketing Algorithm}
\label{section:bucketing} The algorithm is parameterized by a
function $k : \N\to \N$. For a sample of size $m$, the algorithm
splits the interval $[0,1]$ into $k(m)$ equal length subintervals
(\emph{buckets}).  Given an input sample $S$, it counts the number
of sample points lying in each bucket and outputs the mid-point of
the bucket with fewest sample points. In case of ties, it picks the
rightmost bucket. We denote this algorithm by $B_k$. As it turns
out, there exists a choice of $k(m)$ which makes the algorithm $B_k$
consistent for $\mathcal{F}_1$.

\begin{theorem} \label{theorem:bucketing-consistency}
If the number of buckets $k(m) = o(\sqrt{m})$ and $k(m) \to \infty$
as $m \to \infty$, then the bucketing algorithm $B_k$ is consistent
for $\mathcal{F}_1$.
\end{theorem}

\begin{proof}
Fix $f\in \mathcal{F}_1$, assume $f$ has a unique minimizer $x^*$.
Fix $\epsilon,\delta> 0$. Let $U = (x^* - \epsilon/2, x^* +
\epsilon/2)$ be an neighbourhood of the unique minimizer $x^*$. The
set $[0,1] \setminus U$ is compact and hence there exists $\alpha :=
\min f([0,1] \setminus U)$. Since $x^*$ is the unique minimizer of
$f$, $\alpha > f(x^*)$ and hence $\eta:=\alpha - f(x^*)$ is
positive. Thus, we can pick a neighbourhood $V$ of $x^*$, $V \subset
U$, such that for all $x \in V$, $f(x) < \alpha - \eta/2$.

The assumptions on growth of $k(m)$ imply that there exists $m_0$ such that for all $m \ge m_0$
\begin{align}
1/k(m) & < |V|/2 \\
2 \sqrt{\frac{\ln(1/\delta)}{m}} & < \frac{\eta}{2 k(m)} \label{equation:delta_big}
\end{align}

Fix any $m \ge m_0$. Divide $[0,1]$ into $k(m)$ buckets
each of length $1/k(m)$. For any bucket $I$, $I \cap U = \emptyset$,
\begin{equation}
\mu(I) \ge \frac{\alpha}{k(m)} \label{equation:lowerboundmass} \; .
\end{equation}
Since $1/k(m) < |V|/2$ there exists a bucket $J$ such that $J \subseteq V$. Furthermore,
\begin{equation}
\mu(J) \le \frac{\alpha - \eta/2}{k(m)} \label{equation:upperboundmass} \; .
\end{equation}

For a bucket $I$, we denote by $|I \cap S|$ the number of sample
points in the bucket $I$. From the well known Vapnik-Chervonenkis
bounds~\cite{AB99}, we have that with probability at least
$1-\delta$ over i.i.d. draws of sample $S$ of size $m$, for any
bucket $I$,
\begin{equation}
\left| \frac{|I \cap S|}{m} - \mu(I) \right| \le \sqrt{\frac{\ln(1/\delta)}{m}} \; .
\label{equation:vc-bound}
\end{equation}

Fix any sample $S$ satisfying the inequality (\ref{equation:vc-bound}) . For any bucket $I$, $I \cap U=\emptyset$,
\begin{align*}
\frac{|J \cap S|}{m}
& \le \mu(J) + \sqrt{\frac{\ln(1/\delta)}{m}} & \text{by (\ref{equation:vc-bound})} \\
& \le \frac{\alpha - \eta/2}{k(m)} + \sqrt{\frac{\ln(1/\delta)}{m}} & \text{by (\ref{equation:upperboundmass})} \\
& < \frac{\alpha}{k(m)} - 2\sqrt{\frac{\ln(1/\delta)}{m}} + \sqrt{\frac{\ln(1/\delta)}{m}} & \text{by (\ref{equation:delta_big})} \\
& \le \mu(I) - \sqrt{\frac{\ln(1/\delta)}{m}} & \text{by (\ref{equation:lowerboundmass})} \\
& \le \frac{|I \cap S|}{m} & \text{by (\ref{equation:vc-bound})}
\end{align*}
Since $|J \cap S| > |I \cap S|$, the algorithm $B_k$ must not output
the mid-point of any bucket $I$ for which $I \cap U = \emptyset$.
Henceforth, the algorithm's output, $B_k(S)$, is the mid-point of an
bucket $I$ which intersects $U$. Thus the estimate $B_k(S)$ differs
from $x^*$ by at most the sum of the radius of the neighbourhood $U$
and the radius of the bucket. Since the length of a bucket is $1/k <
|V|/2$ and $V \subset U$, the sum of the radii is
$$
|U|/2 + |V|/4 < \frac{3}{4}|U| < \epsilon \; .
$$

Combining all the above, we have that for any $\epsilon, \delta > 0$
there exists $m_0$ such that for any $m \ge m_0$, with probability
at least $1-\delta$ over the draw of an i.i.d. sample $S$ of size
$m$, $|B_k(S) - x^*| < \epsilon$. This is the same as saying that
$B_k$ is consistent for $f$. \qed
\end{proof}

Note that in the above proof we cannot replace the condition $k(m) =
o(\sqrt{m})$ with $k(m) = O(\sqrt{m})$ since Vapnik-Chervonenkis
bounds do not allow us to detect $O(1/\sqrt{m})$-difference between
probability masses of two buckets.

The following theorems shows that if there are too many buckets the bucketing
algorithm is not consistent anymore.

\begin{theorem}
\label{theorem:bucketing-rate} If the number of buckets $k(m) =
\omega(m/\log m)$, then $B_k$ is not consistent for $\mathcal{F}_1$.
\end{theorem}

To prove the theorem we need a proposition of the following lemma dealing with the classical coupon collector problem.

\begin{lemma}[The Coupon Collector Problem \cite{RandomizedAlgo1995}]
\label{lemma:coupon-collector}
Let the random variable $X$ denote the number of trials for collecting each of the $n$ types of coupons.
Then for any constant $c \in \R$, and $m = n \ln n + c n$,
$$
\lim_{n \to \infty} \Pr[X > m] = 1 - e^{-e^{-c}} \; .
$$
\end{lemma}

\begin{proof}[of Theorem~\ref{theorem:bucketing-rate}]
Consider the following density $f$ on $[0,1]$,
$$
f(x) =
\begin{cases}
(4-16x)/3 & \text{if $x \in [0, \frac{1}{4}]$} \\
(16x-4)/3 & \text{if $x \in ( \frac{1}{4}, \frac{1}{2})$} \\
4/3 & \text{if $x \in [\frac{1}{2}, 1]$} \\
\end{cases}
$$
which attains unique minimum at $x^*=1/4$.

From the assumption on the growth of $k(m)$ for all sufficiently large $m$, $k(m) > 4$ and $k(m) > 8m/\ln m$.
Consider the all buckets lying in the interval $[\frac{1}{2}, 1]$
and denote them by $b_1, b_2, \dots, b_n$. Since the bucket size is less than $1/4$, they cover the interval $[\frac{3}{4},1]$.
Hence their length total length is at least $1/4$ and hence there are
$$
n \ge k(m)/4 > 2m/\ln m
$$ such buckets.

We will show that for $m$ large enough, with probability at least $1/2$, at least one of the buckets $b_1, b_2, \dots, b_n$ receives no sample point.
Since probability masses of $b_1, b_2, \dots, b_n$ are the same, we can think of these buckets
as coupon types we are collecting and the sample points as coupons.
By Lemma~\ref{lemma:coupon-collector}, it suffices to verify, that the number of trials, $m$, is at most $\frac{1}{2} n \ln n$. Indeed, we have
$$
\frac{1}{2} n \ln n \ge \frac{1}{2} \frac{2m}{\ln m} \ln \left( \frac{2m}{\ln m} \right)
= \frac{m}{\ln m} \left( \ln m + \ln 2 - \ln \ln m \right) \ge m \; ,
$$
where the last inequality follows from that large enough $m$. Now,
Lemma~\ref{lemma:coupon-collector} implies that for sufficiently large $m$,
with probability at least $1/2$, at least one of the buckets $b_1, b_2, \dots,
b_n$ contains no sample point.

If there are empty buckets in $[\frac{1}{2},1]$, the algorithm outputs a point in $[\frac{1}{2},1]$.
Since this happens with probability at least $1/2$ and since $x^* = 1/4$, the algorithm cannot be consistent. \qed
\end{proof}

When the number of buckets $k(m)$ is asymptotically somewhere in
between $\sqrt{m}$ and $m/\ln m$, the bucketing algorithm switches
from being consistent to failing consistency. It remains an open
question to determine where exactly the transition occurs.

\subsection{The Hard-Margin Algorithm}
\label{section:hard-margin}

Let the \emph{hard-margin} algorithm be the function that outputs
the mid-point of the largest interval between the adjacent sample
points.  More formally, given a sample $S$ of size $m$, the
algorithm sorts the sample $S\cup\{0,1\}$ so that $x_0 = 0 \le x_1
\le x_2 \le \dots \le x_m \le 1 = x_{m+1}$ and outputs the midpoint
$(x_i+x_{i+1})/2$ where the index $i$, $0 \le i \le m$, is such that
the gap $[x_i, x_{i+1}]$ is the largest.

Henceforth, the notion \emph{largest gap} refers to the length of
the largest interval between the adjacent points of a sample.

\begin{theorem}
The hard-margin algorithm is consistent for the family
$\mathcal{F}_1$.
\end{theorem}

To prove the theorem we need the following property of the distribution of 
the largest gap between two adjacent elements of $m$ points forming an i.i.d. 
sample from the uniform distribution on $[0,1]$. The statement of which 
we present an (up to our knowledge) new proof has been originally proven 
by L\'{e}vy~\cite{Levy39}. 

\begin{lemma}
\label{lemma:concentrate} Let $L_m$ be the random variable denoting
the largest gap between adjacent points of an i.i.d. sample of size
$m$ from the uniform distribution on $[0,1]$. For any $\epsilon>0$
$$\lim_{m\rightarrow \infty} \Pr\left[L_m\in \left((1-\epsilon)\frac{\ln m}{m},(1+\epsilon)\frac{\ln m}{m}\right)\right]=1.$$
\end{lemma}

\begin{proof}[of Lemma]
Consider the uniform distribution over the unit circle. Suppose we
draw an i.i.d. sample of size $m$ from this distribution. Let $K_m$
denote the size of the largest gap between two adjacent samples. It
is not hard so see that the distribution of $K_m$ is the same as
that of $L_{m-1}$. Furthermore, since
$\frac{\ln(m)/m}{\ln(m+1)/(m+1)} \to 1$, we can thus prove the lemma
with $L_m$ replaced by $K_m$.

Fix $\epsilon>0$.  First, let us show that for $m$ sufficiently large $K_m$ is
with probability $1-o(1)$ above the lower bound $(1-\epsilon)\frac{\ln m}{m}$.
We split the unit circle $b=\frac{m(1-\epsilon)}{\ln m}$ buckets, each of length
$(1-\epsilon)\frac{\ln m}{m}$. It follows from Lemma~\ref{lemma:coupon-collector}, that for any constant $\zeta>0$ and an i.i.d. sample of $(1-\zeta)b\ln b$ points
at least one bucket is empty with probability $1-o(1)$. We show that for some $\zeta$, $m\leq (1-\zeta)b\ln b$. The expression on the right side can be
rewritten as
\begin{align*}
(1-\zeta)b\ln b
& = (1-\zeta)(1+\delta)\frac{m}{\ln m}\ln\left((1-\zeta)(1+\delta)\frac{m}{\ln m}\right) \\
&\ge m(1-\zeta)(1+\delta)\left(1-O\left(\frac{\ln\ln m}{\ln m}\right)\right)
\end{align*}
For $\zeta$ sufficiently small and $m$ sufficiently large the last
expression is greater than $m$, yielding that a sample of $m$ points
misses at least one bucket with probability $1-o(1)$. Therefore, the
largest gap $K_m$ is with probability $1-o(1)$ at least
$(1-\epsilon)\frac{\ln m}{m}$.

Next, we show that for $m$ sufficiently large, $K_m$ is with probability $1-o(1)$
below the upper bound $(1+\epsilon)\frac{\ln m}{m}$. We consider $3/\epsilon$ bucketings $\mathcal{B}_1, \mathcal{B}_2, \dots, \mathcal{B}_{3/\epsilon}$.
Each bucketing $\mathcal{B}_i$, $i = \left\{1,2, \dots, (3/\epsilon) \right\}$, is a division of the unit circle into $b=\frac{m}{(1+\epsilon/3)\ln
m}$ equal length buckets; each bucket has length $\ell=(1+\epsilon/3)\frac{\ln m}{m}$. The bucketing $\mathcal{B}_i$
will have its left end-point of the first bucket at position $i(\ell
\epsilon/3)$. The position of the left end-point of the first bucket of a
bucketing is called the \emph{offset} of the bucketing.

We first show that there exists $\zeta > 0$ such that $m \ge (1+\zeta)b \ln b$ for all sufficiently large $m$.
Indeed,
\begin{align*}
(1+\zeta)b \ln b & = (1+\zeta) \frac{m}{(1+\epsilon/3)\ln m} \ln\left( \frac{m}{(1+\epsilon/3)\ln m} \right) \\
& \le \frac{1+\zeta}{1+\epsilon/3} m \left(1 - O\left(\frac{\ln \ln m}{\ln m}\right) \right) \; .
\end{align*}
For any $\zeta < \epsilon/3$ and sufficiently large $m$ the last expression is greater than $m$.

The existence of such $\zeta$ and Lemma~\ref{lemma:coupon-collector}
guarantee that for all sufficiently large $m$, for of each bucketing
$\mathcal{B}_i$, with probability $1-o(1)$, each bucket is hit by a
sample point. We now apply union bound and get that, for all
sufficiently large $m$, with probability $1-(3/\epsilon)o(1) = 1 -
o(1)$, for each bucketing $\mathcal{B}_i$, each bucket is hit by at
least one sample point. Consider any sample $S$ such that for each
bucketing, each bucket is hit by at least one point of $S$. Then,
the largest gap in $S$ can not be bigger than the bucket size plus
the difference of offsets between two adjacent bucketings, since
otherwise the largest gap would demonstrate an empty bucket in at
least one of the bucketings. In other, words the largest gap, $K_m$,
is at most
$$K_m \le (\ell \epsilon/3) + \ell = (1+\epsilon/3) \ell = (1+\epsilon/3)^2\frac{\ln m}{m} < (1+\epsilon)\frac{\ln m}{m} \;$$
for any $\epsilon < 1$.

\qed
\end{proof}

\begin{proof}[of the Theorem]
Consider any two disjoint intervals $U, V \subseteq [0,1]$ such that for any $x \in U$ and any $y \in V$,  $\frac{f(x)}{f(y)} < p < 1$
for some $p \in (0,1)$. We claim that with probability $1-o(1)$, the largest gap in $U$ is bigger
than the largest gap in $V$.

If we draw an i.i.d. sample $m$ points from $\mu$, according to the law of large numbers for an arbitrarily small $\chi>0$,
the ratio between the number of points $m_U$ in the interval $U$ and the number of points $m_V$ in the interval $V$
with probability $1-o(1)$ satisfies

\begin{equation}
\label{eq:bound_density}
\frac{m_U}{m_V} \le p(1+\chi)\frac{|U|}{|V|}.
\end{equation}

For a fixed $\chi$, choose a constant $\epsilon>0$ such that $\frac{1-\epsilon}{1+\epsilon} > p+\chi$.

From Lemma~\ref{lemma:concentrate} we show that with probability
$1-o(1)$ the largest gap between adjacent sample points falling into
$U$ is at least $(1-\epsilon)|U|\frac{\ln m_U}{m_U}$. Similarly,
with probability $1-o(1)$ the largest gap between adjacent sample
points falling into $V$ is at most $(1+\epsilon)|V|\frac{\ln
m_V}{m_V}$. From (\ref{eq:bound_density}) it follows that the ratio
of gap sizes with probability $1-o(1)$ is at least
\begin{eqnarray*}
&\ &\frac{(1-\epsilon)|U|\frac{\ln m_U}{m_U}}{(1+\epsilon)|V|\frac{\ln m_V}{m_V}} > \frac{1-\epsilon}{1+\epsilon}\frac{1}{p+\chi}\frac{\ln m_U}{\ln m_V} = (1+\gamma)\frac{\ln m_U}{\ln m_V} \\
&\geq &(1+\gamma)\frac{\ln((p+\chi)\frac{|U|}{|V|}m_V)}{\ln m_V}=(1+\gamma)\left(1+O(1)/{\ln m_V}\right) \to (1+\gamma) \qquad \text{as $m \to \infty$}
\end{eqnarray*}
for a constant $\gamma>0$ such that $1+\gamma\leq \frac{1-\epsilon}{1+\epsilon}\frac{1}{p+\chi}$.
Hence for sufficiently large $m$ with probability $1-o(1)$, the largest gap in $U$ is strictly bigger
than the largest gap in $V$.

Now, we can choose intervals $V_1, V_2$ such that $[0,1] \setminus (V_1 \cup V_2)$ is an arbitrarily small neighbourhood containing $x^*$.
We can pick an even smaller neighbourhood $U$ containing $x^*$ such that for all $x \in U$ and all $y \in V_1 \cup V_2$, $\frac{f(x)}{f(y)} < p < 1$
for some $p \in (0,1)$. Then with probability $1-o(1)$, the largest gap in $U$ is bigger than largest gap in $V_1$ and the largest gap in $V_2$.
\qed
\end{proof}

\section{Learning Linear Cut Separators in High Dimensions}
\label{section:highdim}

%


In this section we consider the problem of learning the  minimum
density homogeneous (i.e. passing through origin) linear cut in
distributions over $\R^d$. Namely, assuming that some unknown
probability distribution generates i.i.d. finite sample of points in
$\R^d$. We wish to process these samples to find the
$(d-1)$-dimensional hyperplane, through the origin of $\R^d$, that
has the lowest probability density with respect to the
sample-generating distribution. In other words, we wish to find how
to cut the space $\R^d$ through the origin in the ``sparsest
direction".\\

 Formally, let $\mathcal{F}_d$ be the family of all
probability distributions over the $\R^d$ that have a continuous
density function. We wish to show that there exists a linear cut
learning algorithm that is consistent for $\mathcal{F}_d$. Note by
Theorem \ref{theorem:non-uniform-convergence}, no algorithm achieves
uniform convergence for $\mathcal{F}_d$ (even for $d=1$).

Define the {\em soft-margin} algorithm with parameter $\gamma:\N \to
\R^+$ as follows. Given a sample $S$ of size $m$, it counts for
every hyperplane, the number of sample points lying within distance
$\gamma:=\gamma(m)$ and outputs the hyperplane with the lowest such
count. In case of the ties, it breaks them arbitrarily. We denote
this algorithm by $H_\gamma$. Formally, for any weight vector
$\mathbf{w} \in \mathcal{S}^{d-1}$ (the unit sphere in $\R^d$) we
consider the ``$\gamma$-strip''
$$
h(\mathbf{w},\gamma) = \{ \mathbf{x} \in \R^d ~:~ |\mathbf{w}^T \mathbf{x}| \le \gamma \}
$$
and count the number of sample points lying in it. We output the weight vector $\mathbf{w}$ for which the number of sample points
in $h(\mathbf{w}, \gamma)$ is the smallest; we break ties arbitrarily.

To fully specify the algorithm, it remains to specify the function
$\gamma(m)$. As it turns out, there is a choice of the function
$\gamma(m)$ which makes the algorithm consistent.

\begin{theorem}
\label{theorem:hyperplanes-consistency} If $\gamma(m) =
\omega(1/\sqrt{m})$ and $\gamma(m) \to 0$ as $m \to \infty$, then
$H_\gamma$ is consistent for $\mathcal{F}_d$.
\end{theorem}

\begin{proof}
The structure of the proof is similar to the proof of
Theorem~\ref{theorem:bucketing-consistency}. However, we will need
more technical tools.

First let's fix $f$. For any weight vector $w \in \mathcal{S}^{d-1}$
and any $\gamma > 0$, we define
$\overoverline{f}_\gamma(\mathbf{w})$ as the $d$-dimensional
integral
$$
\overoverline{f}_{\gamma}(w) := \int_{h(\mathbf{w},\gamma)} f(\mathbf{x}) \; \ud \mathbf{x}
$$
over $\gamma$-strip along $\mathbf{w}$. Note that for any
$\mathbf{w} \in \mathcal{S}^{d-1}$,
$$
\lim_{m \to \infty} \frac{\overoverline{f}_{\gamma(m)}(\mathbf{w})}{\gamma} = \overline{f}(\mathbf{w})
$$
(assuming that $\gamma(m) \to 0$). In other words, the sequence of
functions $\left\{ \overoverline{f}_{\gamma(m)}/\gamma(m)
\right\}_{m=1}^\infty$, $\overoverline{f}/\gamma(m):
\mathcal{S}^{d-1} \to \R_0^+$, converges point-wise to the function
$\overline{f}:\mathcal{S}^{d-1} \to \R_0^+$.

Note that $\overoverline{f}/\gamma(m):\mathcal{S}^{d-1} \to \R_0^+$
is continuous for any $m$, and recall that $\mathcal{S}^{d-1}$ is
compact. Therefore the sequence $\left\{
\overoverline{f}_{\gamma(m)}/\gamma(m) \right\}_{m=1}^\infty$
converges uniformly to $\overline{f}$. In other words, for every
$\zeta > 0$ there exists $m_0$ such that for any $m \ge 0$ and any
$\mathbf{w} \in \mathcal{S}^{d-1}$,
$$
\left| \frac{\overoverline{f}_{\gamma(m)}(\mathbf{w})}{\gamma(m)} - \overline{f}(\mathbf{w}) \right| < \zeta \; .
$$

Fix $f$ and $\epsilon, \delta > 0$. Let $U=\{ \mathbf{w} \in
\mathcal{S}^{d-1} ~:~ |\mathbf{w}^T \mathbf{w}^*| > 1 - \epsilon \}$
be the ``$\epsilon$-double-neighbourhood'' of the antipodal pair
$\{\mathbf{w}^*, -\mathbf{w}^* \}$. The set $\mathcal{S}^{d-1}
\setminus U$ is compact and hence $\alpha := \min
\overline{f}(\mathcal{S}^{d-1} \setminus U)$ exists. Since
$\mathbf{w}^*, -\mathbf{w}^*$ are the only minimizers of
$\overline{f}$, $\alpha > \overline{f}(\mathbf{w}^*)$ and hence
$\eta := \alpha - \overline{f}(\mathbf{w}^*)$ is positive.

The assumptions on $\gamma(m)$ imply that there exists $m_0$ such that for all $m \ge m_0$,
\begin{align}
2\sqrt{\frac{d + \ln(1/\delta)}{m}} & < \frac{\eta}{3} \, \gamma(m) \\
\label{equation:half-spaces-uniform-convergence} \left|
\frac{\overoverline{f}_{\gamma(m)}(\mathbf{w})}{\gamma(m)} -
\overline{f}(\mathbf{w}) \right| & < \eta/3 & \text{for all
$\mathbf{w} \in \mathcal{S}^{d-1}$}  \
\end{align}

Fix any $m \ge m_0$. For any $\mathbf{w} \in \mathcal{S}^{d-1}
\setminus U$, we have
\begin{align*}
\frac{\overoverline{f}_{\gamma(m)}(\mathbf{w})}{\gamma(m)}
& > \overline{f}(\mathbf{w}) - \eta/3  & \text{by (\ref{equation:half-spaces-uniform-convergence})} \\
& \ge  \overline{f}(\mathbf{w}^*) + \eta  - \eta/3 & \text{ by choice of $\eta$ and $U$} \\
& =  \overline{f}(\mathbf{w}^*)  + 2\eta/3 \\
& > \frac{\overoverline{f}_{\gamma(m)}(\mathbf{w^*})}{\gamma(m)} - \eta/3 + 2\eta/3 & \text{by (\ref{equation:half-spaces-uniform-convergence})} \\ \\
& = \frac{\overoverline{f}_{\gamma(m)}(\mathbf{w^*})}{\gamma(m)} + \eta/3  \; .
\end{align*}
From the above chain of inequalities, after multiplying by $\gamma(m)$, we have
\begin{equation}
\overoverline{f}_{\gamma(m)}(\mathbf{w}) > \overoverline{f}_{\gamma(m)}(\mathbf{w^*})+ \eta \gamma(m)/3 \; .
\end{equation}

From the well known Vapnik-Chervonenkis bounds~\cite{AB99}, we have
that with probability at least $1 - \delta$ over i.i.d. draws of $S$
of size $m$ we have that for any $\mathbf{w}$,
\begin{equation}
\label{equation:half-spaces-vc-bound}
\left| \frac{|h(\mathbf{w}, \gamma) \cap S|}{m} - \overoverline{f}_{\gamma(m)}(\mathbf{w}) \right| \le  \sqrt{ \frac{d + \ln(1/\delta)}{m}} \; ,
\end{equation}
where $|h(\mathbf{w}, \gamma) \cap S|$ denotes the number of sample points lying in the $\gamma$-strip $h(\mathbf{w}, \gamma)$.

Fix any sample $S$ satisfying the inequality
(\ref{equation:half-spaces-vc-bound}). We have, for any $\mathbf{w}
\in \mathcal{S}^{d-1} \setminus U$,
\begin{align*}
\frac{|h(\mathbf{w}, \gamma) \cap S|}{m}
& \ge \overoverline{f}_{\gamma(m)}(\mathbf{w}) - \sqrt{ \frac{d + \ln(1/\delta)}{m}} \\
& > \overoverline{f}_{\gamma(m)}(\mathbf{w^*}) + \eta \gamma(m)/3 - \sqrt{ \frac{d + \ln(1/\delta)}{m}} \\
& \ge \frac{|h(\mathbf{w^*}, \gamma) \cap S|}{m} - \sqrt{ \frac{d + \ln(1/\delta)}{m}} + \eta \gamma/3 - \sqrt{ \frac{d + \ln(1/\delta)}{m}} \\
& > \frac{|h(\mathbf{w^*}, \gamma) \cap S|}{m}
\end{align*}
Since $|h(\mathbf{w}, \gamma) \cap S| >  |h(\mathbf{w^*}, \gamma)
\cap S|$, the algorithm must not output a weight vector $\mathbf{w}$
lying in $\mathcal{S}^{d-1} \setminus U$. In other words, the
algorithm's output, $H_\gamma(S)$, lies in $U$ i.e. $|H_\gamma(S)^T
\mathbf{w}^*|
> 1 - \epsilon$.

We have proven, that for any $\epsilon, \delta > 0$, there exists
$m_0$ such that for all $m \ge m_0$, if a sample $S$ is drawn i.i.d.
from $f$, then  $|H_\gamma(S)^T \mathbf{w}^*| > 1 - \epsilon$. In
other words, $H_\gamma$ is consistent for $f$. \qed
\end{proof}

\section{The impossibility of Uniform Convergence}
\label{section:no-uniform-convergence}

In this section we show a negative result that roughly says one
cannot hope for an algorithm that can achieve $\epsilon$ accuracy
and $1-\delta$ confidence for sample sizes that only depend on these
parameters and not on properties of the probability measure.

\begin{theorem}
\label{theorem:non-uniform-convergence} No linear cut learning
algorithm is uniformly convergent for $\mathcal{F}_1$ with respect
to any of the distance functions $D_E$, $D_f$ and $D_\mu$.
\end{theorem}

\begin{proof}
For a fixed $\delta>0$ we show that for any $m\in \mathbb{N}$ there
are distributions with density functions $f$ and $g$ such that no
algorithm using a random sample of size at most $m$ drawn from one
of the
 distributions chosen uniformly at random, can identify the distribution
with probability of error less than 1/2 with probability at least
$\delta$ over random choices of a sample.

Since for any $\delta$ and $m$ we find densities $f$ and $g$ such
that with probability more than $(1-\delta)$ the output of the
algorithm is bounded away by $1/4$ from either $1/4$ or $3/4$, for
the family $\mathcal{F}_1$ no algorithm converges uniformly w.r.t.
any distance measure.

Consider two partly linear density functions $f$ and $g$ defined in
$[0,1]$ such that for some $n$, $f$ is linear in the intervals
$[0,\frac{1}{4}-\frac{1}{2n}]$,
$[\frac{1}{4}-\frac{1}{2n},\frac{1}{4}]$,
$[\frac{1}{4},\frac{1}{4}+\frac{1}{2n})]$, and
$[\frac{1}{4}+\frac{1}{2n},1]$, and  satisfies
$$f(0)=f\left(\frac{1}{4}-\frac{1}{2n}\right)=f\left(\frac{1}{4}+\frac{1}{2n}\right)=f(1),\ f\left(\frac{1}{4}\right)=0,$$ and $g_m$ is the reflection of $f_m$
w.r.t. to the centre of the unit interval, i.e. $f(x)=g(1-x)$. The
functions $f$ and $g$ can be simply described as constant functions
anywhere except of a thin $V$-shape around $1/4$ resp. $3/4$ with
the bottom at 0 in each of them. For any $x\notin
[\frac{1}{4}-\frac{1}{2n},\frac{1}{4}+\frac{1}{2n}]\cup
[\frac{3}{4}-\frac{1}{2n},\frac{3}{4}+\frac{1}{2n}]$, $f(x)=g(x)$.
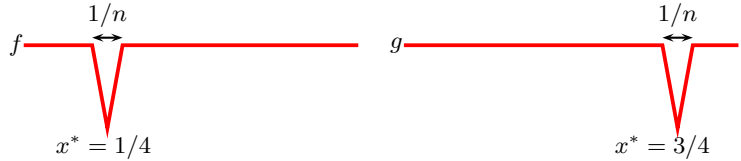
\begin{figure}[htp]
\begin{center}
\begin{pspicture}
\psgrid(0,0.7)(10,1.8)
\psline[linecolor=red,linewidth=0.05](0.1,1.7)(1,1.7)(1.2,0.6)(1.4,1.7)(4.5,1.7)
\psline{<->}(1,1.8)(1.4,1.8) \rput(1.2, 2.1){$1/n$}
\rput(0,1.7){$f$} \rput(1.15,0.35){$\xs=1/4$}

\psline[linecolor=red,linewidth=0.05](5.1,1.7)(8.5,1.7)(8.7,0.6)(8.9,1.7)(9.5,1.7)
\psline{<->}(8.5,1.8)(8.9,1.8) \rput(8.7, 2.1){$1/n$}
\rput(5,1.7){$g$} \rput(8.5,0.35){$\xs=3/4$}

\end{pspicture}
\end{center}
\caption{$f$ is uniform everywhere except a small neighbourhood
around 1/4 where it has a sharp `v' shape. And $g$ is the reflection
of $f$ about $x=1/2$.} \label{fig:uniconv_fig}
\end{figure}

Let us lower-bound the probability that a sample of size $m$ drawn
from $f$ misses the set $U\cup V$ for
$U:=[\frac{1}{4}-\frac{1}{2n},\frac{1}{4}+\frac{1}{2n}]$ and
$V:=[\frac{3}{4}-\frac{1}{2n},\frac{3}{4}+\frac{1}{2n}]$. For any
$x\in U$ and $y\notin U$, $f(x)\leq f(y)$, and furthermore, $f$ is
constant on the set $[0,1]\setminus U$ containing at most the entire
 probability mass 1. Therefore, for $p_f(Z)$ denoting the probability that a point drawn from the distribution with
the density $f$ hits the set $Z$, we have $p_f(U)\leq p_f(V)\leq
\frac{1}{n-1}$, yielding that $p_f(U\cup V)\leq \frac{2}{n-1}$.
Hence, an i.i.d. sample of size $m$ misses $U\cup V$ with
probability at least $(1-2/(n-1))^m\geq (1-\eta)e^{-2m/n}$ for any
constant $\eta>0$ and $n$ sufficiently large. For a proper $\eta$
and $n$ sufficiently large we get $(1-\eta)e^{-2m/n}>1-\delta$. From
the symmetry between $f$ and $g$, a random sample of size $m$ drawn
from $g$ misses $U\cup V$ with the same probability.

We have shown that for any $\delta>0$, $m\in \mathbb{N}$, and for
$n$ sufficiently large, regardless of whether the sample is drawn
from either of the two distributions, it does not intersect $U\cup
V$ with probability more than $1-\delta$. Since in $[0,1]\setminus
(U\cup V)$ both density functions are equal, the probability of
error in the discrimination between $f$ and $g$ conditioned on that
the sample does not intersect $U\cup V$ cannot be less than $1/2$.

\qed

\end{proof}

\section{Conclusions and open questions}
In this paper have presented a novel unsupervised learning problem
that is modest enough to allow learning algorithm with asymptotic
learning guarantees, while being relevant to several central
challenging learning tasks. Our analysis can be viewed as providing
justification to some common semi-supervised learning paradigms,
such as the maximization of margins over the unlabeled sample or the
search for empirically-sparse separating hyperplanes. As far as we
know, our results provide the first performance guarantees for these
paradigms.

From a more general perspective, the paper demonstrates some type of
meaningful information about a data generating probability
distribution that can be reliably learned from finite random samples
of that distribution, in a fully non-parametric model -- without
postulating any prior assumptions about the structure of the data
distribution. As such, the search for a low-density data separating
hyperplane can be viewed as a basic tool for the initial analysis of
unknown data. Analysis that can be carried out in situations where
the learner has no prior knowledge about the data in question and
can only access it via unsupervised random sampling.

Our analysis raises some intriguing open questions. First, note that
while we prove the universal consistency of the `hard-margin'
algorithm for Real data distributions, we do not have a similar
result for higher dimensional data. Since searching for empirical
maximal margins is a common heuristic, it is interesting to resolve
the question of consistency of such algorithms.

Another natural research direction that this work calls for is the
extension of our results to more complex separators. In clustering,
for example, it is common to search for clusters that are separated
by sparse data regions. however, such between-cluster boundaries are
often not linear. Can one provide any reliable algorithm for the
detection of sparse boundaries from finite random samples when these
boundaries belong to a richer family of functions?

Our research has focused on the information complexity of the task.
However, to evaluate the practical usefulness of our proposed
algorithms, one should also carry a computational complexity
analysis of the low-density separation task. We conjecture that the
problem of finding the homogeneous hyperplane with largest margins,
or lowest density around it (with respect to a finite high
dimensional set of points) is NP-hard (when the Euclidean dimension
is considered as part of the input, rather than as a fixed constant
parameter). however, even if this conjecture is true, it will be
interesting to find efficient approximation algorithms for these
problems.\\

\textbf{Acknowledgements.} We would like to thank Noga Alon for a fruitful discussion.


\bibliographystyle{plain}
\bibliography{low-density}

\end{document}